\documentclass[runningheads]{llncs}

\usepackage[numbers]{natbib}
\usepackage[utf8]{inputenc}
\usepackage[T1]{fontenc}
\usepackage{amsmath,amssymb,amsfonts}
\usepackage{graphicx}
\usepackage{textcomp}
\usepackage{url}
\usepackage{booktabs}
\usepackage{multirow}
\usepackage{array}
\usepackage{siunitx}
\usepackage{rotating}
\usepackage{tabularx}
\usepackage{float}
\usepackage{placeins}
\usepackage{algorithm}
\usepackage{algpseudocode}
\usepackage{hyperref}
\usepackage{cleveref}
\usepackage{xcolor}
\usepackage{listings}

\def\tsc#1{\csdef{#1}{\textsc{\lowercase{#1}}\xspace}}
\tsc{WGM}
\tsc{QE}
\begin{document}
\let\WriteBookmarks\relax
\def\floatpagepagefraction{1}
\def\textpagefraction{.001}

\title{AURA: Adaptive Uncertainty-Routed Analysis for Email Threat Detection}

\author{Omran Berjawi\inst{1} \and
Walid fahs\inst{2} \and
Rida Khatoun\inst{1}}

\authorrunning{O. Berjawi et al.}

\institute{Institut Polytechnique de Paris, Télécom Paris, Palaiseau, France
\email{\{omran.berjawi,rida.khatoun\}@telecom-paris.fr}
\and
Islamic University of Lebanon, Faculty of Engineering, Wardanieh, Lebanon
\email{walid.fahs@iul.edu.lb}}

\maketitle

% Here goes the abstract
\begin{abstract}
Email spam and phishing attacks remain a critical security threat. Adversaries increasingly exploit large language models to craft contextually convincing malicious messages, and existing spam detection systems often struggle to keep pace. Generalization across diverse and evolving attack scenarios is limited, which reduces effectiveness once these systems are deployed in practice. This paper introduces Adaptive Uncertainty-Routed Analysis (AURA), a multimodal email threat detection system that analyzes both the content of an email and its embedded URLs. AURA is built around two layers: the first quantifies prediction uncertainty from a URL classifier, and only ambiguous messages are escalated to a fine-tuned transformer encoder for semantic analysis. The system is evaluated on eight heterogeneous training corpora together with two held-out real-world corpora spanning a decade of adversarial campaigns. AURA reaches a macro F1-score of $0.9858$ in-distribution, and on NazPhish-Eval and GuenterTrap-Eval it maintains $0.9502$ and $0.9436$, respectively, which is evidence of robust generalization under genuine distribution shift. 

\end{abstract}

\begin{keywords}
Email spam detection \and 
Phishing detection \and 
Adaptive uncertainty routing \and
Multi-modal classification \and
LoRA fine-tuning \and
Out-of-distribution evaluation \and
Cross-modal signal fusion
\end{keywords}

\section{Introduction}
\label{sec:introduction}
Email has become an indispensable communication medium for individuals, businesses, and government organizations. This makes it a major target for cybercriminals. A substantial share of global email traffic now consists of unsolicited or malicious messages. Phishing remains among the most common initial attack vectors (\cite{ibm2024xforce}), and the volume of phishing attacks continues to reach record highs (\cite{apwg2024}). The problem has only worsened with recent advances in large language models (LLMs). Attackers can now generate highly personalized, context-aware malicious emails. These closely resemble legitimate communications and undercut traditional rule-based and signature-based filtering.

Existing detection approaches share three limitations. Together, they hinder reliable deployment. Many methods still rely on handcrafted features or static rule-based mechanisms. These struggle to keep up with rapidly evolving spam and phishing strategies. A second problem is that learning-based approaches tend to focus on just one signal. They use either the textual semantics of an email or its embedded URL characteristics(~\cite{jamal2024improved,maneriker2021urltran}). This overlooks what the other modality could contribute and narrows detection capability against sophisticated attacks. Finally, the evaluation itself is often the weak point. Methods are commonly tested on balanced or carefully curated benchmark datasets. These do not reflect the distributional shifts and evolving attack patterns found in real email environments. Reported generalization performance, therefore, tends to look better than it actually is. Such protocols miss the degradation that shows up at deployment. Incoming data is temporally and stylistically different from the training distribution because the threat landscape keeps changing(~\cite{janez2022spam}). The result is a gap between benchmark numbers and how reliable these systems actually are once deployed, since rigorous out-of-distribution (OOD) evaluation is rarely performed.

This paper introduces \textbf{AURA}, the \textbf{A}daptive \textbf{U}ncertainty-\textbf{R}outed \textbf{A}nalysis system, a multi-modal email threat detection architecture built to address all three limitations above. Its central mechanism is an Adaptive Uncertainty Router (AUR), which quantifies the prediction uncertainty of a lightweight URL classifier through binary entropy and uses that estimate to decide, per email, what happens next. If the URL-based classification is confident enough, the system reaches an immediate decision. If an email contains no URLs at all, it is sent directly to semantic analysis. And if the signal is genuinely ambiguous, the email is escalated to a fine-tuned DistilBERT encoder that performs deep semantic analysis with cross-modal URL risk injection. The outputs of both analytical stages are then combined through an equal-weight evidential fusion layer to produce the final classification.

Experimentally, AURA achieves a macro F1-score of $0.9858$ on the merged in-distribution test set. On NazPhish-Eval and GuenterTrap-Eval, two corpora derived from real-world spam honeypots and entirely disjoint from the training distribution, it maintains $0.9502$ and $0.9436$.

The primary contributions of this paper are as follows:

\begin{itemize}
\item A multi-modal detection framework that unifies URL structural analysis with deep semantic content analysis, invoking transformer inference selectively for ambiguous cases through an entropy-based routing mechanism.

\item An out-of-distribution evaluation across two temporally diverse real-world corpora, assessing email threat detection robustness under genuine distribution shift across a decade of evolving adversarial campaigns.

\item A modular, deployment-ready architecture in which independently trainable components permit integration with existing email security platforms and straightforward extension with additional detection layers.

\end{itemize}

The remainder of this paper is organized as follows. Section~\ref{sec:related_work} surveys related work and positions AURA within it. Section~\ref{sec:proposed_system} describes the AURA architecture, and Section~\ref{sec:datasets} describes the dataset curation methodology. Section~\ref{sec:experimental_setup} presents the experimental configuration; Section~\ref{sec:results} presents the results. Section~\ref{sec:Deployment} details the deployment and system integration. Section~\ref{sec:discussion} and Section~\ref{sec:limitations_future} interpret the findings, limitations, and future work, respectively. Finally,  Section~\ref{sec:conclusion} concludes the paper.

\section{Related Work}
\label{sec:related_work}
Email spam and phishing detection have been studied for over two decades. The literature organises naturally into three thematic clusters. These are classical and deep learning approaches, transformer and large language model approaches, and URL-based detection. A common weakness runs across all three. Most evaluation protocols rely on in-distribution testing and do not reflect operational deployment conditions.  

\subsection{Classical Machine Learning and Deep Learning Approaches}
\label{subsec:rw_classical_dl}
Classical machine learning classifiers established the foundational vocabulary of spam and phishing detection. These include na\"{i}ve Bayes, support vector machines, logistic regression, and random forests(~\cite{salloum2022systematic,alhuzali2025indepth}). Applied to TF-IDF or bag-of-words email representations, these methods consistently achieved high in-distribution accuracy. Valecha et al.(~\cite{valecha2022persuasion}) enriched the feature space with persuasion-theoretic cues. These included authority, urgency, and social proof. This improved over content-only baselines by $5$--$20\%$ in F-score across multiple phishing corpora. Bountakas and Xenakis(~\cite{bountakas2023helphed}) combined stacking and soft-voting ensemble strategies. They achieved F1\,=\,$0.9942$ on a large imbalanced phishing dataset.

The OOD behaviour of these systems is substantially weaker than their in-distribution figures suggest. Kshirsagar et al.(~\cite{kshirsagar2025meta}) stacked five ML and five deep learning pipelines into a meta-learner. This reached an accuracy of $0.9905$ in-distribution. Spam sensitivity dropped to $0.8970$ on an unseen corpus. This is one of the few explicit OOD evaluations in the spam detection literature. J\'{a}\~{n}ez-Martino et al.(~\cite{janez2022spam}) quantified this degradation directly when na\"{i}ve Bayes and SVM filters trained on standard corpora were evaluated against a strict OOD dataset.

Deep learning approaches replaced hand-crafted features with learned representations. CNN, LSTM, GRU, and their bidirectional variants demonstrated consistent gains over classical classifiers on Enron and SpamAssassin benchmarks. Hybrid architectures that combine BERT embeddings with CNN and GRU layers further improved performance(~\cite{hosseinzadeh2025bert}). Despite these gains, all single-modal deep learning systems share the same structural limitation. They process either text or URL features exclusively. They apply the full inference pipeline to every incoming message regardless of classification difficulty. They are evaluated almost exclusively under in-distribution conditions.

\subsection{Transformer and Large Language Model Approaches}
\label{subsec:rw_transformer_llm}
Pre-trained transformer models advanced the frontier of phishing email detection through fine-tuned contextual representations. This includes BERT and its derivatives(~\cite{devlin2019bert,sanh2019distilbert,liu2019roberta}). Lee et al.(~\cite{lee2020catbert}) proposed CatBERT, a compressed BERT variant fine-tuned for social engineering email detection. "It maintained an 87\% detection rate under adversarial keyword-substitution attacks. Jamal et al.(~\cite{jamal2024improved}) fine-tuned DistilBERT and RoBERTa for joint phishing, spam, and ham classification. They achieved F1\,=\,$0.97$--$0.98$ and showed that parameter-efficient fine-tuning can match full-BERT performance at lower inference cost. Otieno et al.(~\cite{otieno2023bert}) applied fine-tuned BERT to binary phishing classification. They highlighted its sensitivity to adversarial input paraphrasing.

The proliferation of large language models introduced zero-shot and instruction-tuned detection alternatives(~\cite{roumeliotis2024llm,zhang2025benchmarking}). Rojas-Galeano(~\cite{rojas2024zeroshot}) showed that GPT-4 achieves F1\,=\,$0.95$ and Flan-T5 achieves F1\,=\,$0.90$ on SpamAssassin without task-specific training. Koide et al.(~\cite{koide2024chatspam}) proposed the ChatSpamDetector model, which achieved $99.70\%$ accuracy on a custom honeypot corpus. Afane et al.(~\cite{afane2024nazario}) evaluated Gmail, SpamAssassin, Proofpoint, and classical ML classifiers on the Nazario corpus. The results show that performance degraded markedly on LLM-paraphrased variants for all detectors.

Despite strong in-distribution performance, the systems in this cluster share two persistent weaknesses. They process email text exclusively and make no use of URL-structural evidence. They apply full encoder inference uniformly to every incoming message. This design is both informationally incomplete and computationally inefficient at scale(~\cite{salloum2022systematic}).

\subsection{URL-Based Detection}
\label{subsec:rw_url}
URL-based phishing detection is motivated by a simple observation. Malicious links represent a primary attack vector in phishing campaigns. Early work applied classical ML to hand-crafted URL features. These included token entropy, subdomain depth, domain age, and statistics on special characters (\cite{aljofey2020effective,karim2023phishing, sahingoz2019machine}). 

Pre-trained transformer models have since been applied directly to URL strings. Wang et al.(~\cite{wang2023phishbert}) proposed a PhishBERT that was pre-trained on malicious URLs. This achieved a high accuracy on a held-out URL test set. Maneriker et al.(~\cite{maneriker2021urltran}) proposed URLTran. This introduced a URL-specific positional encoding scheme. It outperformed fine-tuned BERT and RoBERTa baselines. Aljofey et al.(~\cite{aljofey2025bertphish}) proposed BERT-PhishFinder. This is a DistilBERT-based URL classifier with optimised tokenisation. It achieved state-of-the-art results on multiple URL benchmarks. Sabir et al.(~\cite{sabir2022reliability}) investigated the robustness of ML-based URL detectors through adversarial perturbation analysis. URL classifiers were shown to be susceptible to inputs crafted to manipulate URL entropy and structural features. This finding directly motivates the entropy-based uncertainty quantification in AURA's Adaptive Uncertainty Router. A fundamental limitation shared by all URL-only methods is their inability to detect social engineering attacks. Phishing content can reside entirely in the message body with no anomalous URL characteristics. AURA addresses this threat category through conditional semantic analysis.

\subsection{Positioning of AURA}
\label{subsec:rw_positioning}
The preceding literature reveals a consistent structural pattern. Existing systems either operate on a single modality, apply deep analysis unconditionally to every incoming message, or evaluate exclusively under in-distribution conditions. Multi-modal architectures combining URL and text signals have been proposed. These include the hierarchical attention model and the late-fusion approach of Zhang et al.~\cite{zhang2022latefusion}. Neither incorporates selective computation allocation nor uncertainty-guided routing. The OOD evaluations of J\'{a}\~{n}ez-Martino et al. ~\cite{janez2022spam}) and Kshirsagar et al. ~\cite{kshirsagar2025meta} confirm that the gap between in-distribution accuracy and real-world robustness is real. AURA addresses all three gaps at once. The entropy-based AUR selectively escalates ambiguous inputs and forces URL-absent emails to the DistilBERT-LoRA encoder. URL and semantic signals are fused adaptively. Performance is validated on two out-of-distribution corpora entirely disjoint from the training distribution.

\section{Proposed System}
\label{sec:proposed_system}
\begin{figure*}[!t]
\centering
\includegraphics[width=\textwidth]{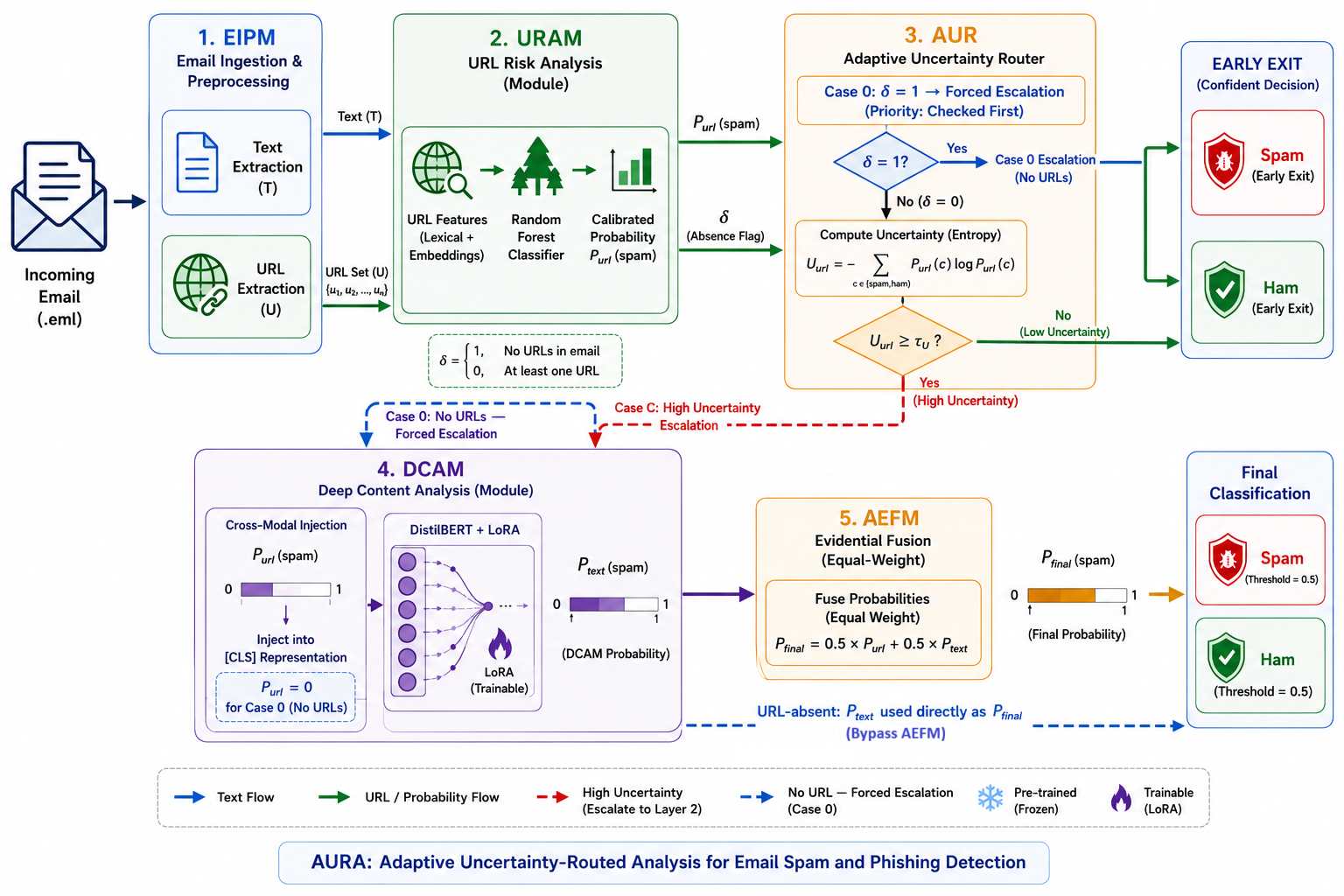}
\caption{Overview of the proposed AURA framework.}
\label{fig:aura_architecture}
\end{figure*}

This section presents the AURA system, which is composed of five modular components as illustrated in Figure~\ref{fig:aura_architecture}. The system first extracts URL and textual features from each incoming email, then performs lightweight URL-based risk analysis. The AUR evaluates prediction confidence and uncertainty to determine whether deeper semantic analysis is required. Ambiguous emails are further analysed using a fine-tuned DistilBERT encoder, and the outputs of both analytical stages are combined to produce the final classification decision. An overview of all AURA modules, their inputs, outputs, and roles is provided in Table~\ref{tab:module_summary}, and the complete inference procedure is formalised in Algorithm~\ref{alg:aura}.

\subsection{Email Ingestion and Preprocessing Module (EIPM)}
\label{subsec:eipm}
All incoming emails are received by AURA in standard \texttt{.eml} format and processed by the EIPM, which serves as the system boundary and produces the structured intermediate representation consumed by all downstream components. Upon receipt of a raw email $e$, EIPM performs two sequential operations. In the first step, the system decodes the email's MIME structure to extract the subject line and both the plain-text and HTML body components. The resulting textual content is then concatenated in a fixed order and subjected to Unicode normalisation, HTML tag stripping, whitespace normalisation, and truncation to a maximum token length compatible with the downstream encoder model. In the second operation, all hyperlinks contained in the email body and selected header fields are identified using regular-expression-based pattern matching, yielding an ordered list of URL strings $\mathcal{U} = \{u_1, u_2, \ldots, u_k\}$.

The output of EIPM is a structured email representation
$\mathcal{E} = \bigl(\mathcal{U},\; T\bigr)$, where $\mathcal{U}$
denotes the extracted URL set and $T$ denotes the normalised text
sequence. The URL set $\mathcal{U}$ is forwarded to URAM for
first-stage analysis, while the text sequence $T$ is retained for
Conditional forwarding to DCAM.

\subsection{Layer 1: URL Risk Analysis Module (URAM)}
\label{subsec:uram}
URAM constitutes the first analytical stage of AURA. It acts as a confidence early-exit classifier for emails whose threat status is unambiguous, avoiding unnecessary semantic analysis for the majority of messages.

\subsubsection{URL Feature Extraction Pipeline}
\label{subsubsec:url_features}
For each URL $u_i \in \mathcal{U}$, URAM constructs a multi-tier feature representation $\mathbf{x}_i = \bigl[\mathbf{f}^{\text{lex}}_i \;\|\; \mathbf{f}^{\text{emb}}_i\bigr]$.

\begin{itemize}
\item \textit{Lexical features} $\bigl[\mathbf{f}^{\text{lex}}_i\bigr]$: derived from the character-level string representation of the URL and capturing structural patterns commonly associated with malicious links. Specifically, URL length and token entropy are computed to quantify structural complexity and randomness. Special character statistics are also extracted, including counts of hyphens, underscores, at-symbols, percent-encoded sequences, digits, and other symbols commonly associated with obfuscated or adversarially crafted URLs. A binary feature indicating HTTPS usage is incorporated to capture whether the secure transport protocol is employed.

\item \textit{Embedding features} $\bigl[\mathbf{f}^{\text{emb}}_i\bigr]$: produced by encoding the full URL string using a SentenceTransformer model based on the MiniLM-L6 architecture into a fixed-dimensional dense vector $\mathbf{f}^{\text{emb}}_i \in \mathbb{R}^{384}$. This learned representation captures distributional regularities in URL patterns that lexical features alone may miss.
    
\end{itemize}

When an email contains multiple URLs ($|\mathcal{U}| > 1$), per-URL feature vectors are aggregated via element-wise maximum pooling: $\mathbf{x}_{\text{url}} = \max_{i=1}^{|\mathcal{U}|} \mathbf{x}_i$. For emails containing no URLs ($\mathcal{U} = \emptyset$), URAM substitutes a zero vector for $\mathbf{x}_{\text{url}}$ and sets a binary absence flag $\delta = 1$ to signal to the AUR that no real URL evidence is available for this message.

\subsubsection{Random Forest Classifier}
\label{subsubsec:rf}
URAM employs a Random Forest (RF) classifier trained on the aggregated URL feature vector $\mathbf{x}_{\text{url}}$ to produce a binary prediction over the label space $\mathcal{Y} = \{\text{spam}, \text{ham}\}$. A critical requirement for URAM is that its output probability scores be well-calibrated, that is, that $P_{\text{url}}(\text{spam})$ reflects the true empirical spam frequency rather than an artefact of the classifier's internal confidence scaling. URAM therefore applies Platt scaling as a post-hoc calibration step, fitting a sigmoid function over the validation set to correct these artefacts and produce a reliable probability estimate. The resulting calibrated probability $P_{\text{url}}(\text{spam}) \in [0, 1]$ represents the degree of confidence that the email is malicious based on its URL content and is subsequently consumed by both the AUR and, conditionally, the AEFM.

\subsection{Adaptive Uncertainty Router (AUR)}
\label{subsec:aur}
The AUR constitutes the architectural core of AURA. It implements a principled routing policy that determines, on a per-email basis, whether (i) the Layer~1 prediction is sufficiently reliable to serve as the final classification decision, (ii) the email should bypass URL-based assessment entirely due to the absence of URL evidence, or (iii) deeper semantic analysis is warranted.

\subsubsection{Entropy-Based Uncertainty Quantification}
\label{subsubsec:entropy}
The calibrated probability $P_{\text{url}}(\text{spam})$ produced by URAM conveys the classifier's estimate of spam likelihood, but a high or low probability value alone does not fully characterise the reliability of that estimate. To address this, the AUR quantifies prediction uncertainty through binary entropy as defined in Equation~\eqref{eq:entropy}:

\begin{equation}
U_{\text{url}} = -P_{\text{url}}(\text{spam})\log_2 P_{\text{url}}(\text{spam}) - (1 - P_{\text{url}}(\text{spam}))\log_2(1 - P_{\text{url}}(\text{spam}))
\label{eq:entropy}
\end{equation}

where $U_{\text{url}} \in [0, 1]$ denotes the prediction uncertainty, with $U_{\text{url}} = 0$ indicating a fully deterministic prediction (i.e., $P_{\text{url}} \in \{0, 1\}$) and $U_{\text{url}} = 1$ indicating maximum uncertainty.

\subsubsection{Conditional Routing Policy}
\label{subsubsec:routing}
Based on the URL absence flag $\delta$, the calibrated probability $P_{\text{url}}(\text{spam})$, and the uncertainty score $U_{\text{url}}$, the AUR applies a four-case conditional routing policy. Let $\tau_{\text{spam}}$, $\tau_{\text{ham}}$, and $\tau_U$ denote the spam confidence threshold, ham confidence threshold, and uncertainty threshold, respectively.

\begin{itemize}
\item \textbf{Case~0 --- URL Absent Forced Escalation:} When URAM signals that no URLs were found in the email ($\delta = 1$), the AUR bypasses all probability-based routing and forces the email directly to DCAM. Case~0 takes precedence over all other cases. When triggered, the DCAM output serves as the sole classification signal, and the AEFM fusion stage is bypassed.

          \begin{equation}
              \text{Route}(e) = \textsc{Escalate}
              \quad \text{if} \quad \delta = 1
              \label{eq:case_0}
          \end{equation}

\item \textbf{Case~A --- Early Spam Exit:} When the URL classifier produces a high spam probability accompanied by low uncertainty, the email is immediately classified as spam without invoking the semantic analysis stage:

          \begin{equation}
              \text{Route}(e) = \textsc{Spam}
              \quad \text{if} \quad
              P_{\text{url}}(\text{spam}) > \tau_{\text{spam}}
              \;\text{ and }\;
              U_{\text{url}} < \tau_U
              \label{eq:case_a}
          \end{equation}

\item \textbf{Case~B --- Early Ham Exit:} Symmetrically, when the URL classifier produces a very low spam probability with low uncertainty, the email is classified as legitimate without semantic analysis:

          \begin{equation}
              \text{Route}(e) = \textsc{Ham}
              \quad \text{if} \quad
              P_{\text{url}}(\text{spam}) < \tau_{\text{ham}}
              \;\text{ and }\;
              U_{\text{url}} < \tau_U
              \label{eq:case_b}
          \end{equation}

\item \textbf{Case~C --- Ambiguous Escalation:} When the uncertainty score meets or exceeds the uncertainty threshold, the prediction is deemed insufficiently reliable and the email is escalated to DCAM for deep semantic analysis:

          \begin{equation}
              \text{Route}(e) = \textsc{Escalate}
              \quad \text{if} \quad U_{\text{url}} \geq \tau_U
              \label{eq:case_c}
          \end{equation}
          Case~C takes precedence over Cases~A and~B whenever
          $U_{\text{url}} \geq \tau_U$, regardless of
          $P_{\text{url}}(\text{spam})$.
\end{itemize}

\subsubsection{Threshold Optimisation via Grid Search}
\label{subsubsec:threshold}
The three routing thresholds $\tau_{\text{spam}}$, $\tau_{\text{ham}}$, and $\tau_U$ are not fixed a priori but are selected through systematic optimisation over the validation set. A grid search is conducted over the Cartesian product of candidate values: $\tau_{\text{spam}} \in [0.5, 0.95]$, $\tau_{\text{ham}} \in [0.05, 0.5]$, $\tau_U \in [0.1, 0.9]$, with a step size of $0.05$ for each parameter. For each candidate threshold combination $(\tau_{\text{spam}}, \tau_{\text{ham}}, \tau_U)$, the macro-averaged F1-score is computed over the validation set to evaluate classification quality. The optimal threshold triple is selected accordingly.

\subsection{Layer 2: Deep Content Analysis Module (DCAM)}
\label{subsec:dcam}
DCAM constitutes the second analytical stage of AURA and is invoked for emails escalated by the AUR under either Case~0 or Case~C. It performs deep semantic analysis of email text content, augmented where available by the URL-derived risk signal from Layer~1. DCAM adopts DistilBERT~\cite{sanh2019distilbert} as its backbone encoder, selected for its demonstrated parameter efficiency and inference speed advantage over BERT. DistilBERT is fine-tuned using Low-Rank Adaptation (LoRA) ~\cite{hu2022lora}), which freezes the pre-trained transformer weights and injects trainable low-rank decomposition matrices into selected attention weight matrices.

The normalised text sequence $T$ produced by EIPM is tokenised and truncated to a maximum length of $256$ tokens. The resulting token sequence is processed by the DistilBERT encoder, and the \texttt{[CLS]} token representation $\mathbf{h}_{\text{cls}}$ extracted from the final hidden layer is concatenated with the scalar URL risk score $P_{\text{url}}(\text{spam})$ from URAM before the classification head %(Equation~\eqref{eq:cross_modal}):
$\mathbf{h}_{\text{fused}} =
    \bigl[\mathbf{h}_{\text{cls}} \;\|\; P_{\text{url}}(\text{spam})\bigr]$.
The fused representation $\mathbf{h}_{\text{fused}}$ is passed through a two-layer classification head consisting of a linear projection with ReLU activation, followed by dropout regularisation and a final linear layer with softmax activation, producing the calibrated probability $P_{\text{text}}(\text{spam}) \in [0, 1]$.

\subsection{Adaptive Evidential Fusion Module (AEFM)}
\label{subsec:aefm}
When an email is escalated to Layer~2 under Case~C, both the URL classifier and the semantic encoder produce independent probability estimates. AEFM combines these two estimates into a single final spam probability by assigning equal weight to URL-structural evidence and semantic text evidence:

\begin{equation}
    P_{\text{final}}(\text{spam}) =
    0.5 \cdot P_{\text{url}}(\text{spam}) +
    0.5 \cdot P_{\text{text}}(\text{spam})
    \label{eq:fusion}
\end{equation}

This design reflects the complementary and informationally distinct nature of the two signals. For emails escalated under Case~0, no URL evidence exists, and AEFM is bypassed entirely; the DCAM output $P_{\text{text}}(\text{spam})$ serves directly as $P_{\text{final}}(\text{spam})$. The final binary classification decision is obtained by thresholding $P_{\text{final}}(\text{spam})$ against a fixed decision boundary of $0.5$, labelling the email as spam if $P_{\text{final}}(\text{spam}) \geq 0.5$ and as ham otherwise.

\begin{table}[H]
\centering
\caption{Summary of AURA module components, inputs, outputs, and design roles.}
\label{tab:module_summary}
\renewcommand{\arraystretch}{1.35}
\begin{tabularx}{\linewidth}{@{}
  p{2.0cm}
  p{1.1cm}
  p{2.2cm}
  p{2.2cm}
  p{2.8cm}
  @{}}
\toprule
\textbf{Module} & \textbf{Abbr.} & \textbf{Input} &
\textbf{Output} & \textbf{Primary Role} \\
\midrule
Email Ingestion \& Preprocessing
  & EIPM & Raw \texttt{.eml} message
  & $(\mathcal{U},\, T)$
  & Parse and structure email into modality streams \\
URL Risk Analysis
  & URAM & URL set $\mathcal{U}$
  & $P_{\text{url}}$,\; $\delta$
  & Lightweight URL threat scoring and absence detection \\
Adaptive Uncertainty Router
  & AUR  & $P_{\text{url}}$,\; $U_{\text{url}}$,\; $\delta$
  & Routing decision (0/A/B/C)
  & Entropy-guided and absence-aware computation allocation \\
Deep Content Analysis
  & DCAM & $T$,\; $P_{\text{url}}$
  & $P_{\text{text}}$
  & Semantic classification with cross-modal input \\
Adaptive Evidential Fusion
  & AEFM & $P_{\text{url}}$,\; $P_{\text{text}}$
  & $\hat{y}$,\; $P_{\text{final}}$
  & Equal-weight multi-modal decision fusion \\
\bottomrule
\end{tabularx}
\end{table}

\begin{algorithm}[ht]
\caption{AURA: Adaptive Uncertainty-Routed Analysis System}
\label{alg:aura}
\begin{algorithmic}[1]
\Require Raw email $e$; trained URAM $\mathcal{F}_1$;
         trained DCAM $\mathcal{F}_2$;
         optimal thresholds $\tau^*_{\text{spam}},\,
         \tau^*_{\text{ham}},\, \tau^*_U$
\Ensure  Binary label $\hat{y} \in
         \{\textsc{Spam}, \textsc{Ham}\}$;
         confidence score $P_{\text{final}}$

\State $(\mathcal{U},\, T) \leftarrow \textsc{EIPM}(e)$
    \Comment{Parse email into URL set and normalised text}
\State $(P_{\text{url}},\, \delta) \leftarrow
       \mathcal{F}_1(\mathcal{U})$
    \Comment{Layer~1: calibrated URL risk scoring}
\If{$\delta = 1$}
    \Comment{Case~0: no URLs found}
    \State $P_{\text{text}} \leftarrow \mathcal{F}_2(T,\, 0)$
        \Comment{DCAM with zero URL risk input}
    \State $P_{\text{final}} \leftarrow P_{\text{text}}$
    \State $\hat{y} \leftarrow \textsc{Spam}$
           \textbf{if} $P_{\text{final}} \geq 0.5$
           \textbf{else} $\textsc{Ham}$
    \State \Return $\hat{y}$,\; $P_{\text{final}}$
\EndIf
\State $U_{\text{url}} \leftarrow
       {-P_{\text{url}}\log_2 P_{\text{url}}
       - (1-P_{\text{url}})\log_2(1-P_{\text{url}})}$
    \Comment{Entropy-based uncertainty}
\If{$P_{\text{url}} > \tau^*_{\text{spam}}$
    \textbf{and} $U_{\text{url}} < \tau^*_U$}
    \State \Return $\hat{y} \leftarrow \textsc{Spam}$,\;
           $P_{\text{final}} \leftarrow P_{\text{url}}$
    \Comment{Case~A: early spam exit}
\ElsIf{$P_{\text{url}} < \tau^*_{\text{ham}}$
       \textbf{and} $U_{\text{url}} < \tau^*_U$}
    \State \Return $\hat{y} \leftarrow \textsc{Ham}$,\;
           $P_{\text{final}} \leftarrow 1 - P_{\text{url}}$
    \Comment{Case~B: early ham exit}
\Else
    \State $P_{\text{text}} \leftarrow
           \mathcal{F}_2(T,\, P_{\text{url}})$
        \Comment{Case~C: Layer~2 semantic analysis}
    \State $P_{\text{final}} \leftarrow
           0.5 \cdot P_{\text{url}} +
           0.5 \cdot P_{\text{text}}$
        \Comment{Equal-weight evidential fusion}
    \State $\hat{y} \leftarrow \textsc{Spam}$
           \textbf{if} $P_{\text{final}} \geq 0.5$
           \textbf{else} $\textsc{Ham}$
    \State \Return $\hat{y}$,\; $P_{\text{final}}$
\EndIf
\end{algorithmic}
\end{algorithm}

\FloatBarrier
\section{Dataset Curation and Collection}
\label{sec:datasets}
This section describes the corpora used to train and evaluate AURA. AURA encompasses two independently trained modules, URAM and DCAM, each trained on a dedicated corpus suited to its input modality. Table~\ref{tab:all_datasets} provides a consolidated overview of all collections, including their type and class distribution.

\subsection{URAM URL Corpus}
URAM is trained on a dedicated corpus of labelled URL records constructed by merging four complementary publicly available URL datasets.

\subsubsection{Sources of URL Collections}

\paragraph{URL-Phish (U1)(~\cite{linh2025feature})}: A feature-engineered phishing URL dataset published in 2025, comprising $100{,}000$ benign samples collected from trusted educational, governmental, and top-ranked domains, and $11{,}660$ phishing samples sourced from the PhishTank repository between November 2024 and September 2025.

\paragraph{LegitPhish (U2)(~\cite{potpelwar2025legitphish})}: A large-scale annotated URL corpus introduced in 2025, comprising $37{,}540$ legitimate entries sourced from Stack Overflow and Wikipedia, and $63{,}678$ phishing URLs obtained from verified threat intelligence feeds including OpenPhish. The dataset is notable for its manually verified labels and the diversity of its legitimate URL sources.

\paragraph{PhishTank URL Corpus (U3)(~\cite{phishtank2025kaggle}))}: A collection of $64{,}280$ real-world phishing URLs drawn from PhishTank.

\paragraph{PhreshPhish (U4)(~\cite{dalton2025phreshphish})}: A large-scale dataset introduced in 2025, comprising $298{,}402$ phishing samples and $367{,}989$ benign HTML--URL pairs collected via a real browser-rendered pipeline to ensure high fidelity of captured page content.

\subsubsection{URL Corpus Integration and Partitioning}
\label{subsubsec:url_partitioning}
The four URL datasets are consolidated into a single unified corpus through two preprocessing operations: label normalisation, whereby phishing and malicious URLs are mapped to $y = 1$ and benign URLs to $y = 0$, and exact-duplicate removal via URL string matching to eliminate redundant entries across sources. The merged corpus is then partitioned using a stratified split strategy applied independently to each source dataset before merging, preserving the class distribution of each source across training ($70\%$), validation ($10\%$), and test ($20\%$) subsets. The resulting partitions are subsequently combined to form the final unified splits.

\subsection{DCAM Email Corpus}
DCAM is trained on a unified multi-domain corpus of eight heterogeneous email collections spanning real-world and synthetically generated messages.

\subsubsection{Sources of Email Collections}
\label{subsubsec:sources}

\paragraph{CSDMC2010 (D1)(~\cite{csdmc2010})}: A spam classification competition corpus comprising raw email messages collected from multiple sources and released as part of the 2010 spam filtering challenge. The corpus contains $2{,}929$ ham and $1{,}378$ spam messages. 

\paragraph{TREC 2005 (D2)(~\cite{cormack2005trec})}: A large-scale corpus released as part of the NIST Text Retrieval Conference Spam Track, collected from a live mail server and adjudicated by human annotators. The corpus contains $39{,}399$ ham and $52{,}790$ spam messages, spanning commercial, social engineering, and bulk unsolicited message types.

\paragraph{TREC 2006 (D3)(~\cite{cormack2006trec})}: The second iteration of the TREC Spam Track evaluation corpus, assembled between April and July 2006 from both user accounts and honeypot addresses deliberately published online to attract spam. The corpus contains $12{,}910$ ham and $24{,}912$ spam messages.

\paragraph{TREC 2007 (D4)(~\cite{cormack2007trec})}: The third iteration of the TREC Spam Track corpus, collected from the same server infrastructure as TREC~2006 over a later temporal window with additional honeypot accounts. The corpus contains $25{,}220$ ham and $50{,}199$ spam messages.

\paragraph{SpamAssassin Public Corpus (D5)(~\cite{spamassassin2006})}: A widely used benchmark corpus assembled by the Apache SpamAssassin project from personal and shared inboxes, categorised into easy ham, hard ham, and spam partitions. The corpus contains $4{,}150$ ham and $1{,}897$ spam messages.

\paragraph{CEAS-08 (D6)(~\cite{segal2008ceas})}: A corpus released as part of the CEAS~2008 Live Spam Filter Challenge, assembled from live email streams during the challenge evaluation period. The corpus contains $17{,}312$ ham and $21{,}842$ spam messages.

\paragraph{PhishFuzzer (D7)(~\cite{toth2025phish})}: An LLM-generated corpus produced by seeding real phishing emails into large language models to generate diverse synthetic phishing variants. The corpus comprises $7{,}700$ ham and $15{,}400$ spam messages.

\paragraph{Spam Genuine (D8)(~\cite{isuranga2023spamgenuine}}: A large-scale synthetically generated corpus containing $100{,}000$ email records designed to simulate the distributional characteristics of real-world messages, balanced with exactly $50{,}000$ ham and $50{,}000$ spam samples.

\subsubsection{Preprocessing and Integration}
\label{subsubsec:preprocessing}
 All eight datasets undergo a two-stage curation pipeline before integration. Each raw email is parsed according to its MIME structure, with the subject and body concatenated into a single normalised text sequence $T$. Duplicate messages are identified and removed via SHA-256 hashing of $T$, non-English text removal, whitespace standardisation, and label normalisation. Each dataset is then independently stratified and split into training ($70\%$), validation ($10\%$), and test ($20\%$) subsets before merging, preserving the spam-to-ham ratio of each source across all partitions. 

\subsection{OOD Evaluation Datasets}
To assess the generalisation capacity of AURA beyond the training distribution, two independent out-of-distribution evaluation corpora are employed. Both corpora are withheld entirely from training and validation and are used solely for hold-out evaluation.

\paragraph{NazPhish-Eval (O1)(~\cite{nazario2005phishing})} This corpus combines the Nazario Phishing Corpus, a long-running collection of real-world phishing emails captured via honeypot addresses between 2015 and 2025, with $3{,}000$ legitimate messages drawn from the Enron Corpus(~\cite{klimt2004enron}). The resulting corpus comprises $3{,}000$ phishing and $3{,}000$ ham messages.

\paragraph{GuenterTrap-Eval (O2)(~\cite{guenter1997spam})} This corpus combines the Bruce Guenter Spam Archive, a continuous spam trap collection harvested from bait addresses configured to attract unsolicited mail, with $39{,}439$ legitimate messages drawn from the Enron Corpus(~\cite{klimt2004enron}). Messages were retrieved covering the period from 2015 to 2025, yielding $39{,}439$ spam-trap messages and $39{,}439$ ham messages.

\begin{table}[H]
\centering
\caption{Summary of all datasets used in the experimental evaluation of AURA.}
\label{tab:all_datasets}
\renewcommand{\arraystretch}{1.3}
\begin{tabular}{lllrrr}
\toprule
\textbf{ID} & \textbf{Dataset} & \textbf{Type} &
\textbf{Ham} & \textbf{Spam} & \textbf{Total} \\
\midrule
\multicolumn{6}{l}{\textit{URL Corpus (URAM Training)}} \\
\midrule
U1    & URL-Phish     & Real                   & 100,000 &  11,660 &  111,660 \\
U2    & LegitPhish    & Real                   &  37,540 &  63,678 &  101,218 \\
U3    & PhishTank     & Real                   &     --- &  64,280 &   64,280 \\
U4    & PhreshPhish   & Real + curated         & 367,989 & 298,402 &  666,391 \\
U\_M  & Merged Cleaned & ---                    & 486,352 & 418,843 &  905,195 \\
\midrule
\multicolumn{6}{l}{\textit{Email Corpus --- In-Distribution (Integrated Training and Evaluation)}} \\
\midrule
D1    & CSDMC2010        & Real --- competition corpus       &   2,929 &   1,378 &   4,307 \\
D2    & TREC 2005        & Real --- live server              &  39,399 &  52,790 &  92,189 \\
D3    & TREC 2006        & Real --- live server / honeypot   &  12,910 &  24,912 &  37,822 \\
D4    & TREC 2007        & Real --- live server / honeypot   &  25,220 &  50,199 &  75,419 \\
D5    & SpamAssassin     & Real --- personal / shared inbox  &   4,150 &   1,897 &   6,047 \\
D6    & CEAS-08          & Real --- live challenge stream    &  17,312 &  21,842 &  39,154 \\
D7    & PhishFuzzer      & LLM-synthetic                    &   7,700 &  15,400 &  23,100 \\
D8    & Spam Genuine     & Synthetic simulation             &  50,000 &  50,000 & 100,000 \\
D\_M  & Merged Cleaned   & ---                              & 156,069 & 185,681 & 341,750  \\
\midrule
\multicolumn{6}{l}{\textit{Email Corpus --- Out-of-Distribution (Hold-Out Evaluation Only)}} \\
\midrule
O1    & NazPhish-Eval    & Real honeypot + Enron ham  &  3,000 &  3,000 &   6,000 \\
O2    & GuenterTrap-Eval & Real spam trap + Enron ham & 39,439 & 39,439 &  78,878 \\
\bottomrule
\end{tabular}
\end{table}

\section{Experimental Setup}
\label{sec:experimental_setup}

\subsection{Implementation Details}
Both modules are trained under an identical stratified split strategy of $70\%$ training, $10\%$ validation, and $20\%$ test, applied independently to each source dataset before merging to preserve class ratios across all partitions. The unified URL corpus yields $633{,}636$ training ($\mathcal{U}_{\text{train}}$), $90{,}520$ validation ($\mathcal{U}_{\text{val}}$), and $181{,}039$ test samples ($\mathcal{U}_{\text{test}}$) . The unified email corpus yields $239{,}225$ training ($\mathcal{D}_{\text{train}}$), $34{,}175$ validation ($\mathcal{D}_{\text{val}}$), and $68{,}350$ test samples ($\mathcal{D}_{\text{test}}$). In both cases, the validation partition is used exclusively for model selection and AUR threshold optimisation, while the test partition is treated as a sealed set accessed only once during final evaluation. The out-of-distribution corpora O1 and O2 are withheld entirely from training and validation, serving solely as hold-out corpora for the generalisation evaluation reported in Section~\ref{sec:results}.

\subsubsection{URAM Configuration}
\label{subsubsec:uram_config}
The URL feature vector $\mathbf{x}_{\text{url}}$ is constructed as described in Section~\ref{subsubsec:url_features}, combining lexical features and a $384$-dimensional SentenceTransformer embedding produced by \texttt{all-MiniLM-L6-v2}. RF was adopted as the URL classifier on the basis of its established strong performance on URL feature representations in prior work and its native probability output suitable for calibration. The RF is configured with $200$ estimators, maximum depth of $10$, minimum samples per split of $5$, minimum samples per leaf of $2$, balanced class weighting, and \texttt{sqrt} maximum features. Probability calibration is applied post-training via Platt scaling on $\mathcal{D}^{u}_{\text{val}}$ to produce the calibrated output $P_{\text{url}}(\text{spam})$ consumed by the AUR.

\subsubsection{DCAM Configuration}
\label{subsubsec:dcam_config}
DCAM is built on the \texttt{distilbert-base-uncased} pre-trained checkpoint, selected for its demonstrated parameter efficiency and inference speed advantage over full BERT. LoRA adapters are injected exclusively into the query (\texttt{q\_lin}) and value (\texttt{v\_lin}) projection matrices of all six attention layers, with all remaining parameters frozen throughout fine-tuning. The LoRA configuration uses rank $r = 16$, scaling factor $\alpha_{\text{LoRA}} = 32$, and dropout $0.1$. Fine-tuning is conducted with a learning rate of $2 \times 10^{-4}$, a batch size of $32$, and the AdamW optimizer, with a maximum of $5$ epochs and early stopping triggered after $2$ consecutive evaluation rounds showing no improvement in macro F1-score on $\mathcal{D}_{\text{val}}$.

\subsubsection{AUR Threshold Configuration}
\label{subsubsec:aur_config}
The routing thresholds $\tau_{\text{spam}}$, $\tau_{\text{ham}}$, and $\tau_U$ are selected through the grid search procedure described in Section~\ref{subsubsec:threshold}, conducted exclusively on $\mathcal{D}_{\text{val}}$. The grid search identified the optimal threshold triple as $\tau^*_{\text{spam}} = 0.80$, $\tau^*_{\text{ham}} = 0.20$, and $\tau^*_U = 0.50$, achieving a macro F1-score of $0.9801$ on $\mathcal{D}_{\text{val}}$. These thresholds are applied uniformly across all evaluation conditions reported in Section~\ref{sec:results}.

\subsection{Evaluation Metrics}
\label{subsec:metrics}
AURA is evaluated under two complementary protocols. Under in-distribution testing, performance is reported on the held-out test partition of each constituent dataset $\mathcal{D}^{\text{test}}_k$ and on the merged test set $\mathcal{D}_{\text{test}}$. Under out-of-distribution testing, AURA is evaluated independently on NazPhish-Eval (O1) and GuenterTrap-Eval (O2) to assess generalisation under genuine distribution shift. Across both protocols, performance is assessed using the following metrics:

\begin{itemize}
\item Precision: the proportion of spam-flagged emails that are truly spam, penalizing false alarms. 

\item Recall: the proportion of actual spam messages correctly detected, penalizing missed threats. 

\item F1-score: the harmonic mean of precision and recall, reported as the macro-averaged value across both classes to account for class imbalance.

\end{itemize}

\subsection{Computational Environment}
\label{subsubsec:hardware}
All experiments are conducted within Kaggle Notebook environments using a single NVIDIA Tesla P100 GPU with $16$~GB of HBM2 memory, $4$ CPU cores, and $30$~GB of host RAM. The implementation relies on the Hugging Face Transformers and PEFT libraries for DCAM fine-tuning and on scikit-learn for URAM classifier training.

\section{Results}
\label{sec:results}

\subsection{In-Distribution Results}
\label{subsec:indistribution}
Table~\ref{tab:indist_results} reports the performance of AURA on the held-out test partition of each of the eight constituent datasets and on the unified merged test set $\mathcal{D}_{\text{test}}$. AURA achieves a macro F1-score of $0.9858$ on the merged test set under the full adaptive routing pipeline. Performance is consistently strong across the six real-world server capture
and competition corpora (D1--D6), with macro F1-scores ranging from $0.9811$
on TREC~2006 (D3) to $0.9979$ on CEAS-08 (D6). TREC~2007 (D4) achieves the
highest precision of any individual dataset at $0.9979$, reflecting the high
proportion of structurally distinct spam in that corpus. CSDMC2010 (D1) and
SpamAssassin (D5) achieve F1-scores of $0.9911$ and $0.9816$, respectively.

\begin{table}[H]
\centering
\caption{In-distribution performance of AURA on the per-dataset test partitions $\mathcal{D}^{\text{test}}_k$ and the unified merged test set $\mathcal{D}_{\text{test}}$.}
\label{tab:indist_results}
\renewcommand{\arraystretch}{1.3}
\begin{tabular}{lrrr}
\toprule
\textbf{Dataset} &
\textbf{Prec} &
\textbf{Rec} &
\textbf{F1} \\
\midrule
CSDMC2010 (D1)    & 0.9960 & 0.9862 & 0.9911 \\
TREC 2005 (D2)    & 0.9928 & 0.9806 & 0.9867 \\
TREC 2006 (D3)    & 0.9828 & 0.9794 & 0.9811 \\
TREC 2007 (D4)    & 0.9979 & 0.9918 & 0.9949 \\
SpamAssassin (D5) & 0.9924 & 0.9711 & 0.9816 \\
CEAS-08 (D6)      & 0.9972 & 0.9987 & 0.9979 \\
PhishFuzzer (D7)  & 0.9784 & 0.9740 & 0.9762 \\
Spam Genuine (D8) & 0.9840 & 0.9810 & 0.9825 \\
\midrule
Merged $\mathcal{D}_{\text{test}}$
                  & 0.9865 & 0.9851 & 0.9858 \\
\bottomrule
\end{tabular}
\end{table}

\subsection{Out-of-Distribution Generalization}
\label{subsec:ood}
Table~\ref{tab:ood_results} reports the performance of AURA on NazPhish-Eval (O1) and GuenterTrap-Eval (O2). Results are reported both per collection year and for the full corpus to enable temporal analysis of detection stability across a decade of evolving adversarial campaigns.

\subsubsection{NazPhish-Eval (O1)}
Across the eleven annual partitions spanning 2015 to 2025, AURA achieves per-year F1-scores ranging from $0.9245$ (2025) to $0.9793$ (2017), with an overall corpus F1-score of $0.9502$. Performance is strongest in the 2017--2021 period, where F1-scores consistently exceed $0.9530$, and exhibits a moderate decline in the 2022--2025 partitions, where scores range from $0.9245$ to $0.9388$. This temporal pattern is consistent with the progressive evolution of phishing campaign characteristics. Specifically, the increasing prevalence of legitimate-looking domains and LLM-assisted message crafting in more recent campaigns.

\subsubsection{GuenterTrap-Eval (O2)}
For the eleven annual partitions spanning 2015 to 2025, AURA achieves per-year F1-scores ranging from $0.9220$ (2025) to $0.9665$ (2018), with an overall corpus F1-score of $0.9436$. The 2019 partition records the lowest F1 of any individual year across both OOD corpora at $0.9123$, suggesting that the spam-trap messages collected in that period exhibit distributional characteristics less well represented in the training corpus. Performance recovers in subsequent years, remaining above $0.9220$ from 2020 onward.  Across both OOD corpora, AURA's overall F1-score differential between in-distribution ($0.9858$) and OOD performance ($0.9436$--$0.9502$) amounts to $3.6$--$4.2$ percentage points, a modest degradation that reflects the domain shift inherent in evaluating on unseen temporal and stylistic distributions.

\begin{table}[H]
\centering
\caption{Out-of-distribution performance of AURA on NazPhish-Eval (O1) and GuenterTrap-Eval (O2), reported per annual partition and for the full corpus.}
\label{tab:ood_results}
\renewcommand{\arraystretch}{1.3}
\begin{tabular}{llrrr}
\toprule
\textbf{Corpus} & \textbf{Year / Partition} &
\textbf{Prec} & \textbf{Rec} & \textbf{F1} \\
\midrule
\multicolumn{5}{l}{\textit{O1 --- NazPhish-Eval
  (Real Phishing + Enron Ham, 2015--2025)}} \\
\midrule
NazPhish-Eval & 2015 & 0.9451 & 0.9744 & 0.9595 \\
              & 2016 & 0.9217 & 0.9781 & 0.9491 \\
              & 2017 & 0.9659 & 0.9932 & 0.9793 \\
              & 2018 & 0.9659 & 0.9772 & 0.9715 \\
              & 2019 & 0.9578 & 0.9809 & 0.9692 \\
              & 2020 & 0.9536 & 0.9528 & 0.9532 \\
              & 2021 & 0.9584 & 0.9556 & 0.9569 \\
              & 2022 & 0.9316 & 0.9462 & 0.9388 \\
              & 2023 & 0.9236 & 0.9467 & 0.9350 \\
              & 2024 & 0.9436 & 0.9096 & 0.9263 \\
              & 2025 & 0.9359 & 0.9133 & 0.9245 \\
\cmidrule{2-5}
              & \textbf{Full corpus}
              & \textbf{0.9458} & \textbf{0.9546}
              & \textbf{0.9502} \\
\midrule
\multicolumn{5}{l}{\textit{O2 --- GuenterTrap-Eval
  (Spam Trap + Enron Ham, 2015--2025)}} \\
\midrule
GuenterTrap-Eval & 2015 & 0.9436 & 0.9516 & 0.9476 \\
                 & 2016 & 0.9369 & 0.9259 & 0.9314 \\
                 & 2017 & 0.9212 & 0.9411 & 0.9310 \\
                 & 2018 & 0.9581 & 0.9751 & 0.9665 \\
                 & 2019 & 0.8969 & 0.9282 & 0.9123 \\
                 & 2020 & 0.9359 & 0.9586 & 0.9471 \\
                 & 2021 & 0.9261 & 0.9504 & 0.9381 \\
                 & 2022 & 0.9335 & 0.9650 & 0.9490 \\
                 & 2023 & 0.9236 & 0.9328 & 0.9282 \\
                 & 2024 & 0.9018 & 0.9484 & 0.9245 \\
                 & 2025 & 0.9136 & 0.9305 & 0.9220 \\
\cmidrule{2-5}
                 & \textbf{Full corpus}
                 & \textbf{0.9364} & \textbf{0.9509}
                 & \textbf{0.9436} \\
\bottomrule
\end{tabular}
\end{table}

 \subsection{Ablation Study}
\label{subsec:ablation}
To isolate the contribution of each analytical layer, two reduced system configurations are evaluated alongside the full AURA system on the merged test set $\mathcal{D}_{\text{test}}$, NazPhish-Eval (O1), and GuenterTrap-Eval (O2):

\begin{itemize}
\item URAM only: the URL classifier alone, applied with a fixed decision threshold of $0.5$ on $P_{\text{url}}(\text{spam})$, with no semantic analysis and no URL-absence handling. 

\item DCAM only: every email is passed unconditionally through the DistilBERT-LoRA encoder without URL risk injection, producing a text-only classification signal.

\end{itemize}

The ablation results are reported in Table~\ref{tab:ablation}. URAM alone produces the weakest performance across all three evaluation conditions, with F1-scores of $0.5487$ on $\mathcal{D}_{\text{test}}$, $0.4591$ on O1, and $0.3339$ on O2. The particularly low score on GuenterTrap-Eval reflects the prevalence of spam-trap messages whose malicious intent is encoded primarily in message content rather than URL structure, rendering URL-only analysis ineffective. The substantially lower score on NazPhish-Eval ($0.4591$) similarly indicates that the phishing campaigns in this corpus frequently exploit legitimate-looking or compromised domains, producing URL entropy scores insufficient for confident classification.

DCAM alone produces substantially stronger performance across all conditions, with F1-scores of $0.9501$ on $\mathcal{D}_{\text{test}}$, $0.9178$ on O1, and $0.9090$ on O2. These results confirm that semantic content analysis provides a more robust and generalisable detection signal than URL structure alone, particularly under distribution shift, where linguistic manipulation patterns remain relatively consistent across phishing campaigns irrespective of URL characteristics.

Full AURA outperforms DCAM alone on all three evaluation conditions, achieving F1-score gains of $+0.0357$ on $\mathcal{D}_{\text{test}}$, $+0.0324$ on O1, and $+0.0346$ on O2. These gains are obtained without any modification to the underlying URAM or DCAM models and are attributable to two complementary mechanisms: the AUR's selective routing, which prevents URL signals from contributing misleading evidence for ambiguous cases, and the equal-weight fusion in AEFM, which leverages partial URL-level evidence for cases where both signals are available.

\begin{table}[H]
\centering
\caption{Ablation study results across the three evaluation corpora.}
\label{tab:ablation}
\renewcommand{\arraystretch}{1.2}
\begin{tabular}{lccc}
\toprule
\textbf{Configuration} &
\textbf{Merged $\mathcal{D}_{\text{test}}$} &
\textbf{O1 NazPhish} &
\textbf{O2 GuenterTrap} \\
\midrule
URAM only       & 0.5487 & 0.4591 & 0.3339 \\
DCAM only       & 0.9501 & 0.9178 & 0.9090 \\
\textbf{AURA}   & \textbf{0.9858} & \textbf{0.9502} & \textbf{0.9436} \\
\bottomrule
\end{tabular}
\end{table}

\subsection{Statistical Significance}
\label{subsec:significance}
To verify that the performance gains of Full AURA over DCAM only, the strongest single-modality baseline, McNemar's test is applied to the per-sample prediction disagreements between the two configurations on each evaluation corpus. McNemar's test is appropriate in this setting because both systems are evaluated on identical test sets, and the test is sensitive to asymmetric disagreements rather than aggregate accuracy. Results are reported in Table~\ref{tab:significance}.

\begin{table}[H]
\centering
\caption{McNemar's test results: Full AURA vs.\ DCAM only.}
\label{tab:significance}
\renewcommand{\arraystretch}{1.2}
\begin{tabular}{lc}
\toprule
\textbf{Evaluation Corpus}  & \textbf{\textit{p}-value} \\
\midrule
Merged $\mathcal{D}_{\text{test}}$  & $0.0152$ \\
O1 --- NazPhish-Eval                & $0.0141$ \\
O2 --- GuenterTrap-Eval             & $0.0075$ \\
\bottomrule
\end{tabular}
\end{table}

The performance gains of Full AURA over DCAM only are statistically significant across all three evaluation corpora ($p < 0.05$). The strongest effect is observed on GuenterTrap-Eval ($p = 0.0075$), where the complementary nature of URL-structural and semantic signals yields the greatest benefit from multi-modal fusion. These results confirm that the improvements attributable to AURA's adaptive routing and fusion mechanisms are reliable and not an artefact of sampling variation.

\subsection{Analysis of Layer Contribution}
\label{subsec:layer_analysis}
The ablation results establish that neither URAM nor DCAM is individually sufficient for robust detection across the full range of evaluated conditions. This subsection further characterises the functional contribution of each layer by analysing the proportion of emails resolved at each stage of the AURA pipeline across the two OOD corpora.

On NazPhish-Eval (O1), $77\%$ of emails are escalated to DCAM, with only $23\%$ resolved at Layer~1. This routing allocation is consistent with the corpus composition: the phishing campaigns in NazPhish-Eval predominantly exploit legitimate-looking or compromised domains, producing URL entropy scores that fall within the uncertainty band $U_{\text{url}} \geq 0.50$ and therefore trigger escalation. The low standalone performance of URAM on this corpus (F1\,=\,$0.4591$) confirms that URL structure alone carries limited discriminative signal for this threat category.

On GuenterTrap-Eval (O2), $55\%$ of emails are resolved at Layer~1, and $45\%$ are escalated to Layer~2. The higher Layer~1 resolution rate reflects the spam-trap origin of this corpus, which predominantly contains bulk unsolicited messages with structurally anomalous URL patterns that fall within the discriminative capacity of the URAM feature pipeline.

This asymmetry in routing allocation across O1 and O2 demonstrates that the AUR adapts its behaviour to the uncertainty characteristics of each incoming distribution rather than applying a fixed escalation strategy. The result is a system that allocates computational resources to the analytical layer best positioned to make a reliable decision for each message, combining the efficiency of URL-based early exit for structurally anomalous spam with the robustness of semantic analysis for sophisticated phishing campaigns that deliberately conceal malicious intent at the URL level.

\section{Deployment and System Integration}
\label{sec:Deployment}
AURA module is integrated with Postfix via the milter interface, in the same manner as SpamAssassin. AURA analyzes email content and embedded URLs to classify emails as spam or legitimate. Hereafter, we analyze the steps required to integrate such a module into Postfix, ensuring compatibility with existing mail server infrastructure. The proposed architecture consists of the following components:

\begin{enumerate}
    \item Postfix mail server: the core SMTP server responsible for email routing.
    \item AURA module.
    \item Integration layer: a middleware component that interfaces between Postfix and AURA.
    \item Database: stores email metadata and model predictions for continuous learning.
\end{enumerate}
The AURA model is deployed as a REST API. An example API endpoint is shown below:

\begin{lstlisting}[caption={AURA REST API request.},captionpos=b]
POST /predict
Content-Type: application/json
{
    "email": {

        "body": "...",
        "embedded URLs": "..."
    }
}
\end{lstlisting}

\begin{lstlisting}[caption={AURA REST API response.},captionpos=b]
{
    "prediction": "spam",
    "confidence": 0.98
}
\end{lstlisting}

The Postfix server usually supports external content filtering via the Milter (mail filter) interface. The integration layer acts as a filter service that processes emails before delivery. The filter service listens on a port and receives the emails from Postfix, extracts headers, body, and metadata, sends the data to AURA for classification, and returns an SMTP response to Postfix: \texttt{250 OK} (accept), \texttt{550 Rejected} (spam), or \texttt{451 Temporary Failure} (quarantine). As for the integration layer, it must parse incoming emails (headers, body, attachments), preprocess the data (e.g., text cleaning, feature extraction), query AURA (via the REST API), and log decisions. We should also configure Postfix to route messages through this filter; for that, we use the Postfix Milter interface (\texttt{smtpd\_milters}) to query AURA before accepting an email.

\section{Discussion}
\label{sec:discussion}

\subsection{Interpretation of Results}
\label{subsec:interpretation}
The central finding of this study is that adaptive, uncertainty-guided routing between a lightweight URL classifier and a deep semantic encoder produces detection performance that is both strong under in-distribution conditions and meaningfully robust under distribution shift. Three design properties of AURA account for this outcome. First, URL-structural and semantic signals are complementary across threat categories: URL analysis efficiently resolves bulk spam with anomalous link patterns. In contrast, semantic analysis covers sophisticated phishing campaigns that deliberately conceal malicious intent at the URL level. Second, entropy-based routing ensures that URL signals contribute to the final decision only when genuinely informative, preventing uncertain predictions from introducing noise into the fusion stage. Third, explicit handling of URL-absent emails, forcing them directly to DCAM, ensures that routing decisions are only trusted when real evidence is available.

\subsection{Generalisation Under Distribution Shift}
\label{subsec:generalisation}
AURA occupies a distinct position relative to prior work, as summarized in Table~\ref{tab:ood_comparison}. AURA's F1-score degradation of $3.6$--$4.2$ percentage points between in-distribution and OOD evaluation is small relative to degradation reported elsewhere in the literature under comparable distribution-shift protocols.  Gmail Spam Filter and SpamAssassin lose $8$--$9$ F1 points when evaluated on LLM-rephrased variants of the Nazario corpus(~\cite{afane2024nazario}). The meta-learner of Kshirsagar et al.(~\cite{kshirsagar2025meta}) loses $24$ AUC points on an unseen 2024 corpus despite near-perfect in-distribution performance. Misra and Rayz(~\cite{misra2022lmsphishing}) report a comparable pattern for fine-tuned BERT: MCC falls from $0.94$ in-domain to $0.72$ on emails drawn from a disjoint domain (IWSPA-AP and Nigerian Fraud corpora). MCC and F1 are not numerically equivalent; this result is reported for qualitative comparison only and is not placed in direct numerical correspondence with the F1 values above.

\begin{table}[H]
\centering
\caption{AURA performance relative to prior systems.}
\label{tab:ood_comparison}
\renewcommand{\arraystretch}{1.2}
\begin{tabular}{lccc}
\toprule
\textbf{System} & \textbf{Metric} & 
\textbf{ID} & \textbf{OOD} \\
\midrule
Gmail Spam Filter(~\cite{afane2024nazario}) 
    & F1 & 0.9679 & 0.8884 \\
SpamAssassin(~\cite{afane2024nazario}) 
    & F1 & 0.9583 & 0.8770 \\
Meta-learner(~\cite{kshirsagar2025meta}) 
    & AUC & 0.9991 & 0.7605 \\
BERT-base(~\cite{misra2022lmsphishing} )
    & MCC & 0.94 & 0.72 \\
\textbf{AURA (proposed)} 
    & F1 & \textbf{0.9858} & 
    \textbf{0.9436--0.9502} \\
\bottomrule
\end{tabular}
\end{table}

The consistent pattern across these systems is that single-modality classifiers, whether rule-based, statistical, or transformer-based, degrade once the evaluation distribution departs from the training distribution. AURA's smaller degradation is consistent with two design choices absent from the compared systems: multi-domain training across eight heterogeneous corpora, which exposes DCAM to a broader range of linguistic manipulation patterns during fine-tuning, and adaptive routing, which prevents an unreliable signal from a single modality from dominating the final decision. The routing allocation data support this: on NazPhish-Eval, where $77\%$ of emails are escalated to DCAM, AURA achieves F1\,=\,$0.9502$ despite URAM alone achieving only $0.4591$ on the same corpus — a $0.4911$ F1 point gap that reflects the extent to which semantic analysis compensates for URL-level evasion.

The most recent partitions (2022--2025) in both OOD corpora show a modest decline relative to earlier years, consistent with the progressive adoption of adversaries' LLM-assisted messages. This suggests that while AURA's semantic representations generalise well across a decade of campaign evolution, the increasing linguistic sophistication of adversarially generated content remains a meaningful challenge for future work.

\subsection{The Role of Adaptive Routing}
\label{subsec:routing_role}
The ablation results show that AURA's gains over DCAM alone come specifically from adaptive routing rather than multi-modal combination alone. Since the DCAM model is identical in both configurations, the $+0.0324$ to $+0.0357$ F1 gains stem entirely from the selective inclusion of URL evidence when informative and its exclusion when not, behaviour confirmed as statistically reliable ($p < 0.05$ across all corpora).

The routing allocation asymmetry between O1 ($77\%$ escalated) and O2 ($45\%$ escalated) shows that the AUR responds to the uncertainty characteristics of each incoming distribution rather than applying a fixed computation policy: on NazPhish-Eval, the system functions effectively as a semantic-only classifier. While on GuenterTrap-Eval, it operates as a true multi-modal detector. This adaptivity distinguishes AURA from fixed-cascade architectures, which apply the same processing pipeline to every message regardless of the information content of individual signals. These findings carry direct implications for deployment. A fixed-cascade system applies full transformer inference to every message, while AURA  reserves this cost for the subset that genuinely requires it, an allocation that scales favourably in high-volume mail environments without sacrificing detection reliability. The remaining gap between in-distribution and OOD performance, together with the unevaluated configurations discussed in Section~\ref{sec:limitations_future}, indicates that this robustness, while substantial, is not unconditional.

\section{Limitations and Future Directions}
\label{sec:limitations_future}
The following limitations of the present study are each paired with a concrete direction for future investigation. 

\subsection{Adversarial Robustness}
AURA's routing component introduces a specific attack surface: an adversary aware of the routing policy could craft structurally benign-looking URLs to produce low uncertainty scores and trigger an early ham exit, bypassing DCAM entirely. Semantically obfuscated phishing messages engineered to evade the DistilBERT encoder represent a complementary threat to the Layer~2 component. As adversaries increasingly exploit the same language models used in detection pipelines, robustness evaluation becomes essential. Future work should assess AURA's resilience to both attack categories and explore adversarial training and certified robustness techniques within the LoRA fine-tuning regime.

\subsection{Language Coverage}
Phishing content varies considerably across languages and regions. All constituent datasets comprise English-language content exclusively, and generalisation of AURA's URL feature pipeline and DistilBERT-LoRA encoder to non-English campaigns cannot be assumed. Future work should extend AURA to multilingual deployment through a multilingual DistilBERT backbone and the construction of multilingual URL training corpora spanning diverse language families.

\subsection{OOD Evaluation Scope}
The two OOD corpora employed cover honeypot and spam-trap collection modalities spanning a decade of real-world campaigns. However, AURA's performance on corporate mail server captures, mobile phishing messages, and social media-based social engineering attacks remains unestablished and may expose limitations not apparent in the current evaluation. Future work should expand the OOD evaluation protocol to encompass these modalities and investigate the applicability of AURA's adaptive routing framework to cross-domain settings, including SMS phishing and instant messaging-based social engineering.

\section{Conclusion}
\label{sec:conclusion}
This paper introduced AURA, an adaptive uncertainty-routed multi-modal email threat detection system. The Adaptive Uncertainty Router provides a principled mechanism for selectively invoking semantic analysis. It is used only when URL signals are genuinely uncertain. It also bypasses URL-based assessment entirely when no URL evidence is present. This prevents silent misclassification of text-only phishing emails. AURA demonstrates robust generalization under in-distribution and out-of-distribution shifts, in which it achieved a macro F1-score of $0.9858$ in-distribution and $0.9502$ and $0.9436$ on the NazPhish-Eval and GuenterTrap-Eval corpora.  The ablation results confirmed that URL and semantic signals are complementary. The AUR escalated $77\%$ of NazPhish-Eval and $45\%$ of GuenterTrap-Eval emails to DCAM. This reflects adaptive routing behaviour across distinct threat distributions.

\bibliographystyle{splncs04}
\bibliography{references}

\end{document}